\documentclass[]{bytedance_seed}
\usepackage{wrapfig}
\usepackage{minted}
\usepackage{bbding}

\usepackage{minitoc}
\usepackage{url}
\usepackage{amssymb}
\usepackage[utf8]{inputenc}
\usepackage{microtype}
\usepackage{booktabs}
\usepackage{pifont} 
\usepackage{multirow}
\usepackage{makecell}
\usepackage{paralist}
\usepackage{xspace}
\usepackage{color}
\usepackage{xcolor}
\usepackage{colortbl}
\usepackage{adjustbox}
\usepackage{hyperref} 
\usepackage[edges]{forest}
\usepackage{tikz} 
\usepackage{caption}
\usepackage{amsfonts}
\usepackage[toc,page,header]{appendix}
\usepackage[utf8]{inputenc}
\usepackage{bbm}
\usepackage{bm}
\usepackage{enumitem}
\usepackage{algorithmicx}
\usepackage{algorithm}
\usepackage{algpseudocode}

\definecolor{seedc}{RGB}{7, 92, 173}

\DeclareMathOperator*{\argmin}{arg\,min}
\DeclareMathOperator*{\mano}{MANOLayer}

\DeclareMathOperator*{\fk}{FK}

\definecolor{wandb-red}{RGB}{240, 67, 79}
\definecolor{wandb-blue}{RGB}{83, 138, 229}
\definecolor{wandb-green}{RGB}{155, 190, 80}
\definecolor{wandb-cyan}{RGB}{91,197, 219}
\definecolor{wandb-brown}{RGB}{172, 111, 80}
\definecolor{wandb-purple}{RGB}{220, 75, 220}

\newcommand{\name}[1]{}
\newcommand{\hardware}[1]{}

\renewcommand{\paragraph}[1]{\vspace{0.1em}\noindent\textbf{#1}}

\newcommand{\greencheck}{\multicolumn{1}{c}{\textcolor{green}{\CheckmarkBold}}}

\newcommand{\redcheck}{\multicolumn{1}{c}{\textcolor{red}{\CheckmarkBold}}}
\newcommand{\redcross}{\multicolumn{1}{c}{\textcolor{red}{\XSolidBrush}}}

\title{DemoBot: Efficient Learning of Bimanual Manipulation with Dexterous Hands From Third-Person Human Videos}

\author[]{ByteDance Seed}

\contribution[]{
Full Author List in Contributions}

\abstract{
Achieving general-purpose embodiment AI requires robots to acquire diverse manipulation skills efficiently and at scale. However, current data-driven paradigms are fundamentally constrained by the scarcity of high-quality robot data, as collecting demonstrations via teleoperation is costly and unscalable. Learning directly from the massive volume of internet-scale human videos offers a promising solution to this data bottleneck, yet it presents a new set of challenges: the significant \textit{embodiment gap} between human and robot hands, and the \textit{modality gap} where visual data lacks the critical physical dynamics information required for task execution. Consequently, direct mimicking of human video remains ineffective and infeasible for complex tasks.

To address these challenges and bridge the gap between visual observation and physical execution, we propose \textbf{DemoBot}, a novel framework that learns long-horizon, bimanual dexterous manipulation skills from a single, unannotated RGB-D video. The core insight of DemoBot is to utilize the extracted human trajectory not as a strict imitation target, but as a flexible \textit{motion prior} for guidance. Firstly, the data processing module extracts structured motion trajectories of both hands and objects from raw video data. These trajectories serve as motion priors for a novel reinforcement learning (RL) pipeline that learns to refine them through contact-rich interactions, thereby eliminating the need to learn from scratch. To further address the challenging of learning long-horizon manipulation skills, we introduce: (1) Temporal-segment based RL to enforce temporal alignment of the current state with demonstrations; (2) Success-Gated Reset strategy to balance the refinement of readily acquired skills and the exploration of subsequent task stages; and (3) Event-Driven Reward curriculum with adaptive thresholding to guide the RL learning of high-precision manipulation. The novel data processing and RL framework successfully achieved long-horizon synchronous and asynchronous bimanual assembly tasks, offering a scalable approach for direct skill acquisition from human videos. Experiments demonstrate that DemoBot can efficiently master contact-rich bimanual skills from a single video demonstration, offering a scalable path toward learning from visual human demonstrations.
}

\date{12 December, 2025}
\correspondence{\email{yucheng.xu@bytedance.com}}
\checkdata[Project Page]{\url{https://demobot-seed.github.io/}}

\begin{document}
\maketitle

\section{Introduction}
\label{sec:intro}

The ability to learn complex manipulation skills directly from observing humans is an essential capability for enabling Embodied Artificial General Intelligence at scale. A true generalist robot should be able to learn new abilities, not from months of time-consuming teleoperation--a process that fundamentally lacks scalability. A more promising path towards generalization is to enable robots to learn from the massive internet-scale of human videos~\cite{mccarthy2025towards}. However, a foundational challenge for such a scalable learning system is the need for efficient data processing and effective skill learning. For a robot to learn skills from massive videos, a core competitive strength is being able to acquire a new skill robustly and efficiently from one single video demonstration. This single-shot learning capability -- similar to an apprentice watching a master craftsperson once -- is one of the most critical building blocks~\cite{mccarthy2025towards}. Achieving this could unlock the potential to leverage vast internet-scale datasets of human activities and create a new generation of general-purpose robots. However, directly translating visual demonstrations into successful robot skills, especially for bimanual dexterous tasks, remains an open challenge in research to date.

A primary obstacle to overcome is to bridge the significant gap between the human demonstrator and the robot agent. Existing approaches for learning from visual demonstration often rely on specialized hardware, such as teleoperation systems with VR~\cite{ding2024bunny, cheng2024open, triantafyllidis2023hybrid} or wearable devices~\cite{mao2023learning, mao2024dexskills, tao2025dexwild, shaw2024bimanual, zhang2025doglove}, or motion capture systems~\cite{wang2024dexcap, chen2024object, li2025maniptrans, zhou2025you}, to obtain motion data. Learning directly from a single, unannotated RGB-D video -- a setup that is far more scalable and accessible -- is not trivial. The data lacks action labels and is inherently noisy, object and hand occlusions often occur, and a fundamental embodiment mismatch exists between the human hand and a robot end-effector.

Even with the extracted 3D motion trajectories from visual demonstrations, another major challenge arises --~learning a robust policy from imperfect data. Long-horizon manipulation requires capturing not only the accurate kinematic motion, but also the physical dynamics. Conventional Imitation Learning (IL) methods~\cite{ding2024bunny, mao2023learning,  mao2024dexskills, tao2025dexwild, shaw2024bimanual, wang2024dexcap, zhang2025doglove, zhou2025you} are often designed for high-quality, kinematically accurate, and physically feasible demonstrations. This assumption, however, does not apply to data derived from visual demonstrations, which lack critical information about physical dynamics due to modality gap. While reinforcement learning (RL) methods~\cite{chen2024object, chen2023bi} are capable of learning contact-rich skills through extensive interaction with the physical world, they present their own set of difficulties. Learning long-horizon bimanual dexterous manipulation skills via RL from scratch is notoriously sample-inefficient and difficult due to the high-dimensional state-action spaces associated with multi-fingered hands. Besides, further challenges also include temporal misalignment with the demonstration~\cite{huey2025imitation}, effective exploration in vast state-action spaces, and assigning credit over long time scales~\cite{gupta2019relay, ni2023transformers}.

\begin{figure*}[!t]
\centering
\includegraphics[trim=8mm 0mm 8mm 0mm,clip,width=\linewidth]{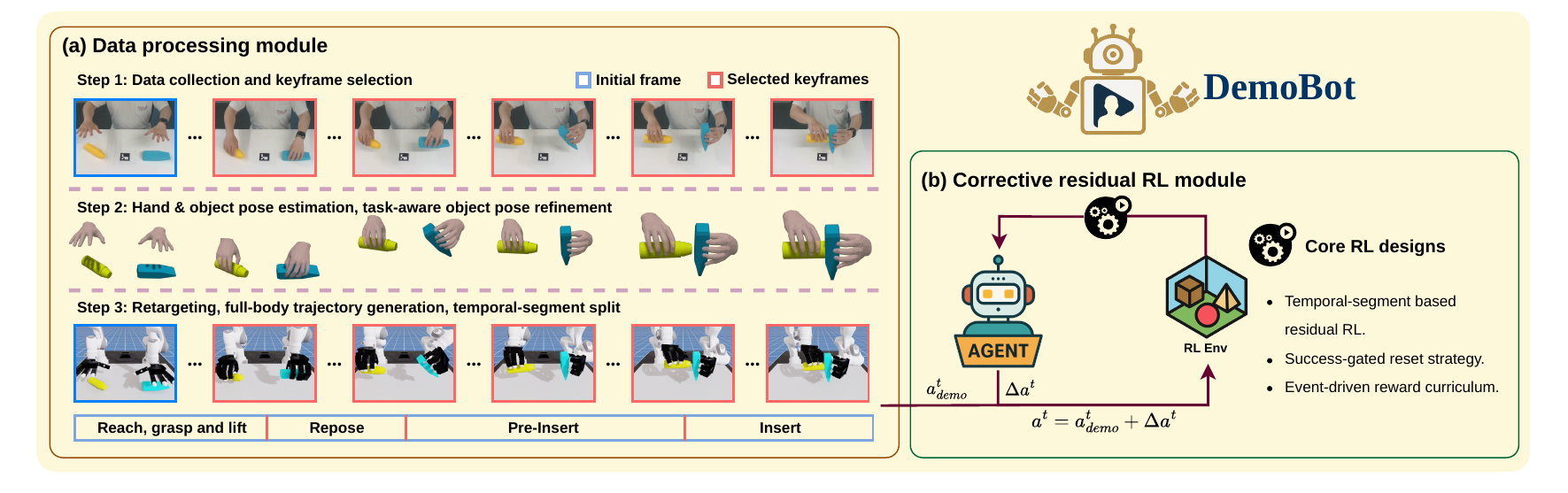}
\caption{The DemoBot framework for learning bimanual skills from a single visual demonstration. (a) The \textbf{Data Processing Module} converts a raw RGB-D video into structured motion priors in three steps: (1) A human demonstration is recorded and manually segmented with keyframes. (2) Hand and object estimators produce 3D hand and object poses, which are then refined using task-aware optimization. (3) The refined human motion is retargeted to the robot, generating a full-body trajectory that is split into meaningful temporal segments based on the keyframes. (b) The \textbf{Corrective Residual RL Module} then uses these segments as motion priors. The RL agent learns a corrective policy that outputs a residual action, $\Delta a$,  which is added to the motion priors, $a = a_{demo} + \Delta a$, allowing the robot to master the contact-rich physical dynamics absent from the original visual data and complete the task. 
}
\label{fig:main}
\end{figure*}

To address these challenges, we proposed a novel framework, \textbf{DemoBot}, that offers the ability to learn bimanual dexterous manipulation skills from third-person human videos -- as few as one single unannotated RGB-D video demonstration. This framework consists of a robust data processing module and a novel residual reinforcement learning pipeline. The data processing module extracts both object and human hand motions from RGB-D video and then further maps them to full-body robot action trajectories and a set of object sub-goals to achieve. The residual RL pipeline solves two main challenges: (1) Learning robot skills from a single noisy and imperfect demonstration; (2) Solving long-horizon and complex bimanual manipulation tasks. Firstly, the core idea for learning from single demonstration is to treat the extracted imperfect motion trajectories from visual demonstration as a plausible yet imperfect \textit{motion prior} as guidance, rather than an enforced target to be mimicked. The RL pipeline learns a corrective residual policy that locally refines the motion prior to account for the physical dynamics,  which is missing from the original visual demonstration, by interacting with the physical simulation, and completing the task. Then, to deal with the challenges emerging from the long-horizon and complex bimanual manipulation tasks, we introduce three main novel designs into the RL pipeline: (1) \textbf{Temporal-segment based RL} paradigm for mitigating the temporal misalignment between collected long-horizon demonstration and the current RL state by splitting the entire task into a sequence of segments. Then, the RL agent is trained to complete the goal of each segment sequentially. The use of temporal segments allows the RL agent to learn corrective residual for the motion prior of the current stage independently, avoiding the misleading information from the other stages; (2) \textbf{Success-gated reset strategy} for balancing the retention of early skills with the deep exploration of later ones by randomly resetting the failed environments to their last success terminal state. This ensures that the collect learning samples are allocated across the entire task. (3) \textbf{Event-driven reward curriculum} for a smooth learning signal by 
employing a combination of dense rewards and sparse bonus and further decomposing them into event-related terms (e.g. reaching, lifting, reposing, etc). These terms are controlled and activated when specific condition are met. Additionally, we designed a curriculum learning strategy to gradually anneal the threshold for the sparse bonus, according to the training progress, to gradually improve the precision of manipulation. The proposed RL pipeline combines the macromotion guidance from the visual human demonstration with the contact-rich exploration of RL as well as overcoming typical challenges in long-horizon complex bimanual manipulation tasks. 

The main contributions of the proposed framework are:
\begin{itemize}[leftmargin=*]
    \setlength{\itemsep}{1pt}
    \setlength{\parskip}{0pt}
    \setlength{\parsep}{0pt}

    \item A novel and robust video processing pipeline that optimizes the extracted 3D hand-object motion priors using a MANO-based representation of human hand with task-related refinements for object pose, transferring the unannotated 2D video into high-quality 3D motion priors suited for robot learning -- processing a 15s-video using 4 minutes of compute on a single GPU.
    
    \item A suite of novel reinforcement learning techniques designed for demo-augmented, long-horizon tasks: a \textbf{Temporal-segment based RL} paradigm to mitigate temporal misalignment, a \textbf{Success-gated reset} strategy to enable deep exploration, and an \textbf{Event-driven reward curriculum} to learn high-precision dexterous skills.
    
    \item This learning framework is the first of its kind, to the best of our knowledge, which can efficiently learn long-horizon, bimanual dexterous manipulation skills from a single visual human demonstration. %
\end{itemize}

We validate the effectiveness of DemoBot through a set of challenging long-horizon bimanual manipulation tasks that require precise coordination and contact management. Our extensive experiments demonstrate that the proposed residual learning paradigm successfully bridges the modality and embodiment gaps inherent in visual demonstrations, and mitigates the challenges in long-horizon skill learning. The system is shown to master complex skills where standard behavioral cloning and reinforcement learning baselines struggle, proving that sub-optimal visual data can serve as an effective motion prior when paired with a corrective residual policy.

Ultimately, by enabling robust skill acquisition from a single, unannotated RGB-D video without the need for expensive teleoperation hardware or massive datasets, this work represents a significant step toward scalable robot learning. DemoBot offers a practical pathway for general-purpose robots to expand their skill repertoires by leveraging the massive, open-ended library of human video data available on the internet, moving us closer to the goal of Embodied Artificial General Intelligence.

\section{Related Work}
In this section, we present the review of related works, which are focused on two main area: (1) data collection for learning from demonstration; (2) Demonstration-augmented reinforcement learning.  
\subsection{Data Collection for Learning from Demonstration}
\begin{wraptable}{r}{0.5\linewidth} 
  \centering
  \vspace{-3mm} 
  \small
  \setlength{\tabcolsep}{3pt} 
  \caption{Comparison with recent Learning from Demonstration (LfD) works.}
  \label{tab:summary}
  \begin{tabular}{l|llll}
    \hline
                   & \textbf{Dexterous} & \textbf{Bimanual} & \textbf{LfV} & \textbf{1-shot} \\ \hline
    \multicolumn{1}{c|}{RobotTube\cite{xiong2022robotube}} &\redcross &\greencheck &\greencheck &\redcross        \\
    \multicolumn{1}{c|}{DexMV\cite{qin2021dexmv}} &\greencheck &\redcross &\greencheck &\redcross        \\
    \multicolumn{1}{c|}{YOTO\cite{zhou2025you}} &\redcross &\greencheck &\greencheck &\greencheck        \\
    \multicolumn{1}{c|}{DexCap\cite{wang2024dexcap}} &\greencheck &\greencheck &\redcross &\redcross      \\
    \multicolumn{1}{c|}{ManipTrans\cite{li2025maniptrans}} &\greencheck &\greencheck &\redcross &\redcross  \\ 
    \multicolumn{1}{c|}{The work in~\cite{mao2023learning}} &\redcross &\greencheck &\redcross &\greencheck        \\\hline
    \multicolumn{1}{c|}{\textbf{DemoBot (Ours)}} &\greencheck  &\greencheck &\greencheck &\greencheck  \\\hline
    \end{tabular}
  \vskip -0.2cm 
\end{wraptable}
\label{sec:related_work}
A significant portion of prior learning from demonstration~(LfD) research is designed for high-fidelity data. Teleoperation setups, remote controllers~\cite{ding2024bunny, cheng2024open, triantafyllidis2023hybrid} or wearable devices~\cite{mao2023learning, mao2024dexskills, tao2025dexwild, shaw2024bimanual, zhang2025doglove} provide clean, kinematically-valid action data but require expensive, specialized hardware and significant operator effort. Similarly, motion capture systems~\cite{wang2024dexcap, chen2024object, li2025maniptrans, zhou2025you} with multi-camera arrays can yield precise human motion data, but these are often constrained to laboratory environments and are not easily scalable. In contrast, learning from passive, third-person video is a far more scalable and accessible approach, with the potential to leverage internet-scale data~\cite{qin2023anyteleop, qin2022one, qin2021dexmv, xiong2022robotube}. However, this introduces significant perceptual and morphological challenges. Reconstructing 3D hand-object interactions from a single RGB or RGB-D stream is an ill-posed problem, often resulting in noisy and inaccurate pose estimates. A summary of recent LfD work and their key characteristics is presented in Table.~\ref{tab:summary}. Our work contributes to this area by proposing a robust data pipeline that uses modern, model-based hand estimators~\cite{metro, hamer, wilor} and introduces a novel, task-aware optimization step to refine these imperfect estimates.

\subsection{Demonstration-Augmented Reinforcement Learning}

To overcome the brittleness of Imitation Learning while retaining the sample efficiency of LfD, a major trend has been to combine demonstrations with reinforcement learning~\cite{triantafyllidis2023hybrid, li2025maniptrans, qin2021dexmv, rajeswaran2017learning, hester2018deep, vecerik2017leveraging}. These 
methods leverage demonstrations to guide and accelerate 
RL.
A typical strategy
is to encourage the RL agent to mimic the demonstrations~\cite{rajeswaran2017learning, hester2018deep, vecerik2017leveraging}. However, these methods inherently suffer from  temporal misalignment between the current RL state and the demonstrations, especially in long-horizon tasks. 
Other methods~\cite{triantafyllidis2023hybrid, li2025maniptrans, qin2021dexmv} are designed to further extract deep representations from the collected demonstrations. The work in~\cite{triantafyllidis2023hybrid} learns reward function from demonstration with GAIL~\cite{ho2016generative}. The works in~\cite{li2025maniptrans, qin2021dexmv} learn the state-action mapping from the collected demonstration to initialize the RL policy. However, these methods require multiple demonstrations to ensure the reliability of the learned reward function or action mapping, which is not feasible with single demonstration, increasing the difficulty in scaling. 

Our work is built upon the paradigm of Residual Reinforcement Learning~\cite{johannink2019residual, silver2018residual}, where the demonstrated motion serves as a base trajectory, and the RL agent learns corrective residual actions. The final action is the sum of the base action and the learned residual. This structure effectively transforms a difficult, high-dimensional control problem into a more tractable one for learning local corrections. To overcome the temporal misalignment issue in long-horizon task, the entire demonstration trajectory is split into multiple temporal segments and adapt the residual RL to refine each segment sequentially. By combining these new designs, our work can effectively learn a long-horizon manipulation skill from only a single demonstration. 

\section{Methodology}
In this section, we introduce the core designs of our proposed DemoBot in depth. First, we present the data processing module which is used to extract hand-object motion priors from visual human demonstration. Then, we present our novel demo-augmented residual RL module, which is tailored for learning long-horizon complex bimanual manipulation skill from a single human demonstration. 
\label{sec:method}
\subsection{Hand-object motion priors from visual demonstration}
In this section, we present the data processing pipeline aimed at extracting and refining the motion trajectories from visual demonstrations. 

\subsubsection{Human Demonstration Capture}

\begin{figure*}[!t]
\centering
\includegraphics[trim=6mm 0mm 6mm 0mm,clip,width=1\linewidth]{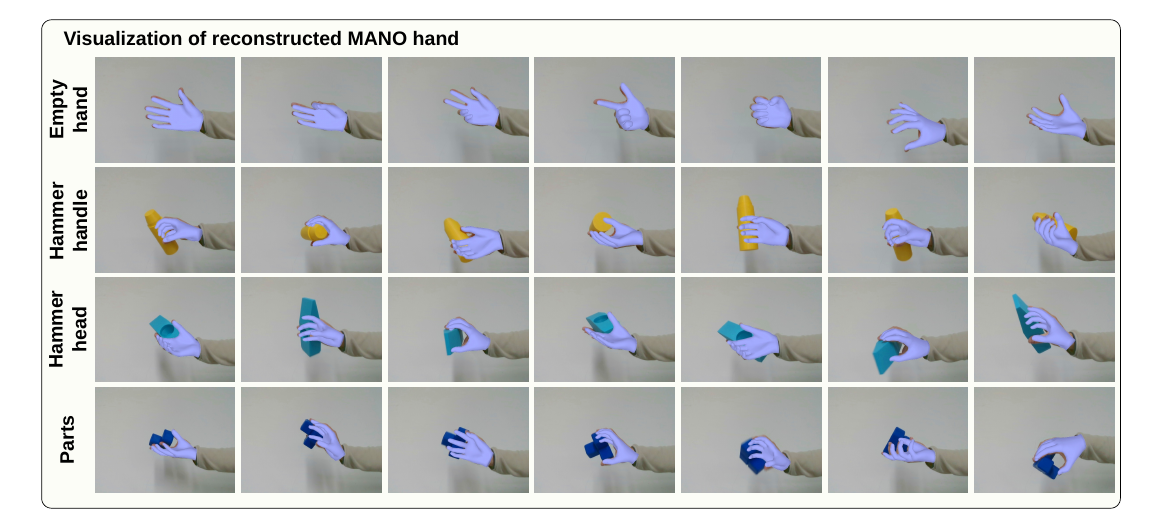}
\caption{Robustness of the MANO-based hand pose estimation under different occlusion cases and hand gestures, with MANO hand model reconstructed from RGB images and visualized by overlaying on the original RGB images.}
\label{fig:mano-vis}
\end{figure*}
\begin{wrapfigure}{r}{0.5\linewidth} 
\centering
\vspace{-5mm}
\includegraphics[width=\linewidth]{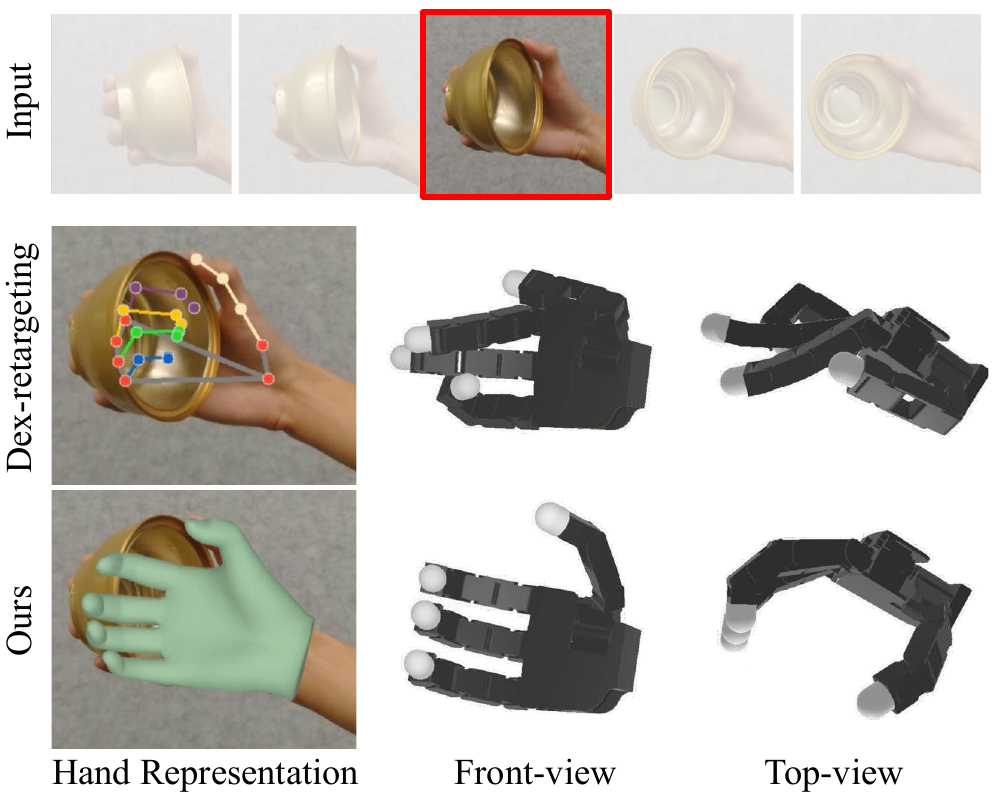}
\vspace{-5mm}
\caption{Advantage of our algorithm preserving the integrity of the full hand pose: comparison of our developed MANO-based retargeting algorithms against the SOTA 2D keypoint-based retargeting.}
\vspace{-8mm}
\label{fig:retarget}
\end{wrapfigure}
Dual-arm human demos are captured using a single depth camera, which provides synchronized RGB-D streams. The camera is pre-calibrated to obtain its intrinsic $\mathbf{K}$ and extrinsic parameters $[\mathbf{R}|\mathbf{t}]$. 
During data recording, an operator manually annotates keyframes corresponding to critical stages of the task (e.g., grasping, lifting, inserting), which can be used to guide subsequent learning stage.

\subsubsection{Hand pose estimation}
\label{subsubsec:data-capture}

To parameterize the motion of both hands, we use the MANO-based hand estimator~\cite{metro, hamer, wilor}
to estimate the 3D hand poses from each RGB frame as MANO~\cite{MANO} parameters $\{\theta, \beta, \mathbf{R}^h, \mathbf{t}^h\}$, where $\theta$ encodes the local hand joint poses, $\beta$ encodes the hand shape, $\mathbf{R}^h \in \mathcal{SO}(3)$ is the global hand orientation, $\mathbf{t}^h \in \mathbb{R}^{3}$ is the global hand translation, together with an estimated camera intrinsic matrix $\hat{\mathbf{K}}$ and detected 2D hand joints $\mathbf{J}^{2d}$. As it can be seen from Figure.~\ref{fig:mano-vis}, the MANO-based hand pose estimator is robust for different scenarios, especially for cases where the human hand is largely occluded by in-hand objects. This process is applied independently to the left and right hands. However, the reconstructed 3D MANO hand is defined under the estimated camera instrinsics $\hat{\mathbf{K}}$, which may differ from the actual calibrated camera's intrinsics $\mathbf{K}$ of our real setup. To resolve this discrepancy, we further align the 3D MANO hand with the observed human hand by back-projecting the 3D MANO hand joints onto the current RGB video frame 
and minimizing the $l$-2 distance between projected MANO hand joints and detected 2D hand joints $\mathbf{J}^{2d}$:
\begin{equation}
\begin{split}
\theta, \beta, \mathbf{R}^h, \mathbf{t}^h & = \argmin_{\theta, \beta, \mathbf{R}^h, \mathbf{t}^h} || \hat{\mathbf{J}}^{2d} - \mathbf{J}^{2d} ||_{2}\\
\text{where} \;\; \hat{\mathbf{J}}^{2d} & = \mathbf{K}\hat{\mathbf{J}}^{3d}, \\ 
\hat{\mathbf{J}}^{3d} &= \mano(\theta, \beta, \mathbf{R}^h, \mathbf{t}^h)
\end{split}
\label{eq:mano-alignment}
\end{equation}
where 
$\mano(*)$ is the differentiable MANO interpreter~\cite{hasson19_obman} which receives MANO parameters and outputs both 2D and 3D hand joints. The above hand pose estimation and alignment results in a sequence of 3D hand poses in MANO parameter space, $\mathcal{T}^{hand} = \{\tau^{hand}_{t}\}_{t=0}^{T}$, where ${\tau^{hand}_{t}} = (\theta_{t}, \beta_{t}, \mathbf{R}^h_{t}, \mathbf{t}^h_{t})$.

\subsubsection{Object pose estimation}
A 2D image segmentor~\cite{cheng2023segment} is firstly applied to the RGB sequence generating object segmentation masks. These masks, along with the RGB-D frames and the 3D object model, are then fed into the off-the-shelf 3D object pose estimator~\cite{wen2024foundationpose} to yield the object's rotation $\mathbf{R}^o \in \mathcal{SO}(3)$ and translation $\mathbf{t}^o \in \mathbb{R}^3$.
Similar to the hand pose estimation, these initial pose estimates can have errors that are detrimental to high-precision tasks like assembly. These errors are reduced by using a task-aware refinement module that optimizes the pose with respect to a task-specific objective function, $f_{task}$ (see Sec~\ref{sec4b}):
$$
\mathbf{R}^o, \mathbf{t}^o = \argmin_{\mathbf{R}^o, \mathbf{t}^o} f_{task}(\mathbf{R}^o, \mathbf{t}^o).
$$
This module is used to compute a sequence of object poses, $\mathcal{T}^{obj} = \{\tau^{obj}_{t}\}_{t=0}^{T}$, where $\tau^{obj}_{t} = (\mathbf{R}^o_{t}, \mathbf{t}^o_{t})$.

\subsubsection{MANO-based Hand to Robot retargeting}
\label{subsec:retarget}

\begin{wrapfigure}{R}{0.5\textwidth}
\vspace{-8mm}
    \begin{minipage}{0.5\textwidth}
        \begin{algorithm}[H]
        \scriptsize
        \caption{Demonstration Replay and Segmentation}
        \label{alg:replay}
        \begin{footnotesize}
        
        \begin{algorithmic}[1]
        \State \textbf{Input:} Processed hand trajectory $\mathcal{T}^{hand} = \{(q^h_t, p^{base}_t)\}_{t=0}^T$
        \State \textbf{Input:} Processed object trajectory $\mathcal{T}^{obj} = \{(\mathbf{R}^o_t, \mathbf{t}^o_t)\}_{t=0}^T$
        \State \textbf{Input:} Set of keyframe indices $\mathcal{K} = \{k_1, k_2, \dots, k_N\}$
        \State \textbf{Output:} A set of temporal segments $\mathcal{S} = \{S_1, S_2, \dots, S_N\}$
        
        \State Initialize replay buffer $\mathcal{B} \leftarrow \emptyset$
        \State Initialize last keyframe index $t_{prev} \leftarrow 0$
        
        \For{$t = 0$ to $T$}
        \State Get current hand base pose $p^{base}_t$
        \State Solve for arm joints: $q^{arm}_t \leftarrow \text{IK}(p^{base}_t)$
        \State Form full-body configuration: $q_t \leftarrow (q^{arm}_t, q^h_t)$
        \State Add current robot state to buffer: $\mathcal{B} \leftarrow \mathcal{B} \cup \{ q_t \}$
        
        \If{$t \in \mathcal{K}$}
        \State Define sub-goal for the stage: $g \leftarrow (\mathbf{R}^o_t, \mathbf{t}^o_t)$
        \State Create stage segment: $S_{new} \leftarrow (\mathcal{B}, g)$
        \State Add segment to set: $\mathcal{S} \leftarrow \mathcal{S} \cup \{S_{new}\}$
        \State Clear the buffer for the next stage: $\mathcal{B} \leftarrow \emptyset$
        \State Update last keyframe index $t_{prev} \leftarrow t$
        \EndIf
        \EndFor
        \end{algorithmic}
        \end{footnotesize}
        \end{algorithm}
    \end{minipage}
\vspace{-10mm}
\end{wrapfigure}

As shown in Figure.~\ref{fig:retarget}, traditional retargeting methods~\cite{qin2023anyteleop, li2019vision, handa2020dexpilot} rely on 2D keypoints, which fail when the hand is occluded by an object. The MANO-based representation is more robust to such occlusions. We retarget the refined 3D hand trajectory $\mathcal{T}^{hand}$ from the previous step to a floating-base robot hand. 
The goal is to estimate the robot hand's joint positions $q_h$ and base pose $p_{base}$ that best mimic the MANO hand's 3D joint locations. We solve this via optimization:
\begin{equation}
\begin{split}
q^{h}, p^{base} & = \argmin_{q, p} || \hat{\mathbf{J}}^{3d} - \mathbf{J}^{3d}_{fk} ||_{2}, \\
\text{where} \;\; \mathbf{J}^{3d}_{fk} &= \fk(q), \\ 
\hat{\mathbf{J}}^{3d} &= \mano(\theta, \beta, \mathbf{R}^h, \mathbf{t}^h)
\end{split}
\label{eq:retargeting}
\end{equation}

$\fk(*)$ is forward kinematics~(FK), $q_{h}$ are the robot joint positions, $p_{base}$ is the robot hand base 3D pose.

\subsubsection{Real-to-Sim Full-Body Trajectory Generation}
\label{subsec:replay}
The retargeting process yields motion only for the robot's hands, but not for the arms. To generate a complete and kinematically-valid trajectory for the full bimanual robot system, we conduct inverse-kinematic-based trajectory generation within the IsaacLab simulator~\cite{mittal2023orbit}. At each timestep $t$ of the demonstration, we have the target robot hand joint positions $q^{h}_{t}$ and the target hand base pose $p^{base}_{t}$ from the retargeting module, along with the refined object pose $(\mathbf{R}^{o}_{t}, \mathbf{t}^{o}_{t})$. To determine the required arm motion, we treat the hand base pose $p^{base}_{t}$ as the end-effector target for the robot arm and solve for the arm joint positions $q^{arm}_{t}$ using Inverse Kinematics (IK). The full-body robot configuration at this timestep is then $q_{t} = \{q^{arm}_{t}, q^{h}_{t}\}$. Then, we leverage the keyframes manually annotated during the data capture (e.g., grasping, lifting, inserting) to split the continous trajectory into \textbf{temporal segments}. This segmented demonstration structure is detailed in Algorithm~\ref{alg:replay}. After replaying the entire demonstration, we are left with a sequence of $N$ temporal segments, $\mathcal{S} = \{S_1, S_2, \dots, S_N\}$, each representing a distinct phase of the manipulation task with a clearly defined goal.

\subsection{Learning bimanual dexterous manipulation skills with residual reinforcement learning}
The data processing pipeline described in the following sections produces a complete, kinematically-valid trajectory $\{q^t\}_{t=0}^T$ for the dual-arm system. These trajectories serve as strong \textit{motion priors} that captures the overall strategy and quasi-dynamically feasible motions of the demonstrated task, they are inherently imperfect and not entirely compliant with physics for being executed directly. 

Here are two primary remaining challenges to be resolved:
\begin{itemize}[leftmargin=*]
    \setlength{\itemsep}{1pt}
    \setlength{\parskip}{0pt}
    \setlength{\parsep}{0pt}
    \item \textbf{Accumulated Estimation Error.} The trajectory is the output of a hierarchical pipeline (pose estimation \(\rightarrow\) alignment \(\rightarrow\) retargeting \(\rightarrow\) IK). Small errors at each stage can accumulate, leading to a final trajectory with kinematic inaccuracies. 
    \item \textbf{Modality Gap.} The visual demonstration, captured via an RGB-D stream, lacks information about actual physical contact and dynamics between hands and objects, which is vital for contact-rich object manipulation tasks. 
\end{itemize}
Consequently, the generated trajectory cannot guarantee that the robot hand will make and maintain correct as well as stable contact with the object, which is critical for successful manipulation.
Instead of discarding this valuable prior, we design our learning module to explicitly leverage it. The core idea is to learn a \textit{corrective residual policy} that refines the imperfect demonstration locally, rather than learning a control policy from scratch. This transforms a high-dimensional, exploration-heavy RL problem into a more tractable one focused on local adjustments. 
To this end, we formulate the task as a \textbf{residual reinforcement learning}.

Define a standard Markov Decision Process (MDP) specified by the tuple \((\mathcal{S}, \mathcal{A}, \mathcal{P}, \mathcal{R}, \gamma)\).
The state at timestep $t$, $s_t = (q_t, \dot{q}_t, s^{obj}_t, s^{task}_t)~\in~\mathcal{S}$ is composed of the robot's joint positions \(q_t\) and velocities \(\dot{q}_t\), state of the manipulated objects \(s^{obj}_t\), as well as other task-related states \(s^{task}_t\). The action \(a_t \in \mathcal{A}\) executed by the simulator is the target joint position for the robot's low-level controller. 
In the residual RL formulation, this action is the sum of a base action from the demonstration and a learned residual action:
$a_t = a^{demo}_t + \Delta a_t$.
Here, \(a^{demo}_t\) is the target joint positions from the processed trajectory at the current timestep, i.e., \(a^{demo}_t = q_{t}\). The residual action \(\Delta a_t\) is the output of a trained neural network policy \(\pi_{\phi}\), which learns to make corrections based on the current state: \(\Delta a_t \sim \pi_{\phi}(\cdot | s_t)\). To explicitly constrain the exploration of the RL agent and make it stay close to the base action trajectory, the predicted residual action \(\Delta a_t\) is clipped to \([-0.25, 0.25]\) radians. The objective of the RL agent is to learn the policy parameters \(\phi\) that maximize the expected cumulative discounted reward. The detailed reward terms are discussed in Section.~\ref{subsubsec:reward}. 

To deal with long-horizon bimanual manipulation tasks, we proposed three novel designs for the integration in the robot learning pipeline: (1) Temporal-segment based Reinfocement Learning, (2) Success-Gated Resets Strategy, (3) Event-Driven Reward curriculum.

\paragraph{\textcolor{seedblue}{Temporal-segment based Reinfocement Learning.}}
A critical challenge in demo-augmented reinforcement learning from a single demonstration is the problem of \textbf{temporal misalignment}. Standard approaches that compel a policy to follow an entire trajectory in a lock-step, time-indexed manner are often brittle~\cite{rajeswaran2017learning}. If the agent's execution speed deviates from the demonstration -- due to dynamics, perturbations, or initial state differences -- it can fall out of sync. This causes the agent to receive wrong base actions corresponding to a different phase of the task, leading to compounding errors and eventual failure.

To circumvent this, we replace full trajectory correction by a temporal-segment based RL paradigm. As detailed in Section~\ref{subsec:replay} and Algorithm~\ref{alg:replay}, the continuous demonstration is segmented into a sequence of meaningful segments \(\mathcal{S} = \{S_1, S_2, \dots, S_N\}\), where each segment \(S_i = (\mathcal{B}_i, g_i)\) consists of a trajectory segment \(\mathcal{B}_i\) and a corresponding state-based sub-goal \(g_i\). The segment id is further included into the state \(s_t\) at each timestep $t$ to explicitly condition the RL policy with the information of the current segment. 

The weights and biases of the final output layer of the policy network \(\pi_{\phi}\) are initialized to zero. These designs inherently facilitates a \textbf{learn-then-optimize} strategy within each stage, which is key to mitigating temporal misalignment. At the beginning of training, the predicted residual action \(\Delta a_t\) is close to zero. Consequently, the agent's initial behavior is to faithfully track the demonstrated states on a step-by-step basis. In this phase, temporal alignment is maintained by design. As training progresses and the agent becomes more competent and confident, the task-oriented reward signal encourages it to transcend the demonstration's specific timing and develop a more generalized, efficient skill for the current stage.

\begin{table*}[!t]
\centering
\caption{Reward terms and physical quantities across environments. } \label{tab:rewards}
\resizebox{\linewidth}{!}{
\begin{tabular}{c|l|c|c|c|ccc}
\hline
\multirow{2}{*}{\textbf{Type}} & \multirow{2}{*}{\textbf{Symbol}} & \multirow{2}{*}{\textbf{Description}}                                                                                               & \multirow{2}{*}{\textbf{Equation}}                 & \multirow{2}{*}{\textbf{scale}} & \multicolumn{3}{c}{\textbf{Stage}}                  \\ \cline{6-8} 
                      &                         &                                                                                                                            &                                           &                        & \textit{reach}       & \textit{grasp and lift} & \textit{goal}        \\ \hline
\multirow{3}{*}{\textbf{Dense reward}} & $r_{reach}$             & \begin{tabular}[c]{@{}c@{}}Distance between hand's end-effector\\ and object's center-of-mass, $d_{h2o}$\end{tabular}      & $1 - tanh(d_{h2o} / 0.3)$                 & 1.0                    & \greencheck & \greencheck    & \greencheck \\ \cline{2-5}
                      & $r_{grasp}$             & \begin{tabular}[c]{@{}c@{}}Distance between hand’s fingertips \\ and object’s cloest surface point, $d_{f2o}$\end{tabular} & $1 - tanh(d_{f2o} / 0.05)$                & 2.0                    &             & \greencheck    & \greencheck \\ \cline{2-5}
                      & $r_{goal}$              & \begin{tabular}[c]{@{}c@{}}Distance between object's keypoints\\ and current goal's keypoints, $d_{goal}$\end{tabular}     & $1 - tanh(d_{goal} / 0.1)$                & 15.0                   &             &                & \greencheck \\ \hline
\multirow{3}{*}{\textbf{Sparse bonus}} & $B_{reach}$             & Object reaching bonus                                                                                                      & $1 \; if \; d_{h2o} < \delta_{reach}, \; else \; 0$     & 50                     & \redcheck   &                &             \\ \cline{2-5}
                      & $B_{lift}$              & Object lifting bonusx                                                                                                       & $1 \; if \; height_{obj} > \delta_{lift}, \; else \; 0$ & 100                    &             & \redcheck      &             \\ \cline{2-5}
                      & $B_{goal}$              & Goal reaching bonusx                                                                                                        & $1 \; if \; d_{goal} < \delta_{goal}, \; else \; 0$     & 1000                   &             &                & \redcheck   \\ \hline
\multirow{2}{*}{\textbf{Task-specific bonus}} & $B_{sync}$                                   & Synchronous goal reaching bonus                                                                                            & \multicolumn{1}{l|}{$1 \; if \; |Time(B_{goal}^{right} - B_{goal}^{left})| < w, \; else \; 0$} & \multicolumn{1}{c|}{2000} &             &                & \redcheck   \\ \cline{2-5}
                      & $B_{switch}$                                 & Switch phase bonus                                                                                                         & \multicolumn{1}{l|}{\textbf{$1 \; if \; current \; stage == switch, \; else \; 0$}}  & \multicolumn{1}{c|}{2000} &             &                & \redcheck   \\ \hline
\end{tabular}}
\end{table*}
\paragraph{\textcolor{seedblue}{Success-Gated Resets Strategy.}}
A novel \textbf{Success-Gated Reset} strategy makes the RL exploration more efficient.
This approach manages the distribution of starting states to balance the retention of early skills with the deep exploration of later ones. When an episode terminates due to failure in the current segment \(S_i\), instead of deterministically resetting to global initial state \(s_0\), we reset probabilistically.
With probability \(p_{init}\), the environment is reset to the global initial state \(s_0\) (as a regularizer against forgetting). With probability \(1 - p_{init}\), it is reset to the successful terminal state of the previous stage, \(s_{i-1}^*\), which then serves as the new starting point. This probabilistic reset strategy can be viewed as a form of implicit curriculum learning: 
\begin{itemize}[leftmargin=*]
    \setlength{\itemsep}{1pt}
    \setlength{\parskip}{0pt}
    \setlength{\parsep}{0pt}
    \item The reset to \(s_{i-1}^*\) provides \textbf{focused training}, allowing the agent to repeatedly practice the specific stage it finds challenging without the ``cost" of re-executing all preceding stages. This dramatically improves sample efficiency for mastering later parts of the task.
    \item The reset to \(s_0\) acts as a \textbf{regularizer against forgetting}, ensuring that the policy continuously rehearses the foundational skills from the beginning, thereby maintaining a robust, holistic solution.
\end{itemize}
The hyperparameter $p_{init}$ controls the balance between exploring the current frontier of the agent's ability and exploiting mastered skills.

\paragraph{\textcolor{seedblue}{Event-Driven Reward curriculum.}}
\label{subsubsec:reward}
 A reward function that combines dense distance-based rewards with sparse event-driven bonuses guides the policy. This hybrid structure provides granular feedback while also signaling the achievement of critical task milestones. The rewards are activated sequentially, creating an implicit curriculum that guides the agent from reaching, to grasping, and finally to manipulating. The total reward at any timestep is a sum of dense and sparse components, as listed in Table.~\ref{tab:rewards}. The green check mark denotes reward terms awarded across the entire stage while the red check mark denotes bonus terms awarded only when a specific stage is completed. Besides general reward components, there are two task-specific bonuses: $B_{sync}$ for encouraging the agent to learn the coordination skill between two arms in the synchronous bimanual task, $B_{switch}$ for encouraging the agent to explore and navigate through the critical ``switch phase''. 
 There are also
 two curriculum strategies: (1) Threshold annealing on the goal reaching threshold~(\(\delta_{goal}\)), which bridges the gap between initial exploration and the high precision~(millimeter-level) contact-rich manipulation; (2) A pre-grasp curriculum that freezes the hand open during the reaching stage. This simplifies the initial stage to arm control only, and then introduces more complex finger control problem when it's needed.

\section{Experimental Validation}
\label{sec:results}
In this section, we first present the experimental setup we used for experiments in simulation and on real-world hardware respectively, then we introduce the implementation details of core modules. After that, we present experimental results in simulation and ablation study to demonstrate the effectiveness of our framework design. Lastly, we report the results on real-world setup, as a proof-of-concept, to demonstrate the sim-to-real transferability of our method.

\begin{figure*}[!t]
\centering
\includegraphics[trim=6mm 0mm 6mm 0mm,clip,width=1\linewidth]{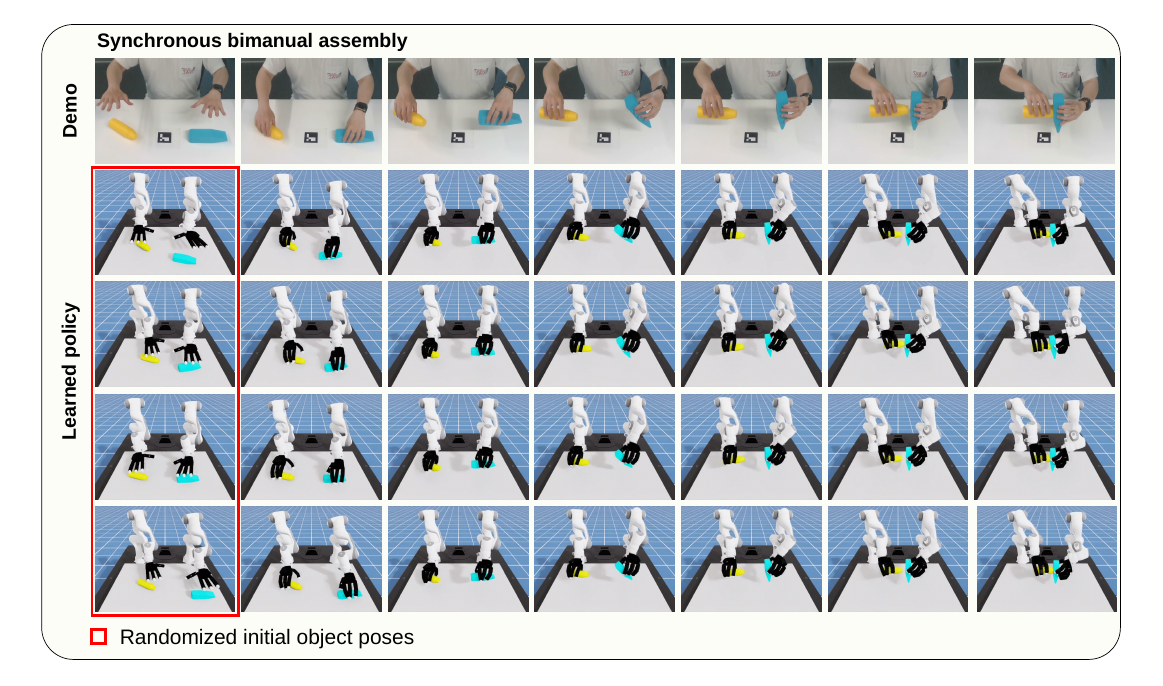}
\caption{Physics-based simulation of bimanual assembly skills learned from human videos for the synchronous assembly task. The top row shows the collected demo and the rest rows show manipulation results with trained policy. We note that although our method is based on single demonstration, applying randomization on the initial poses of objects can still lead to generalization on random initial object poses. }
\label{fig:sim-exp-sync}
\end{figure*}

\begin{figure*}[!t]
\centering
\includegraphics[trim=6mm 0mm 6mm 0mm,clip,width=1\linewidth]{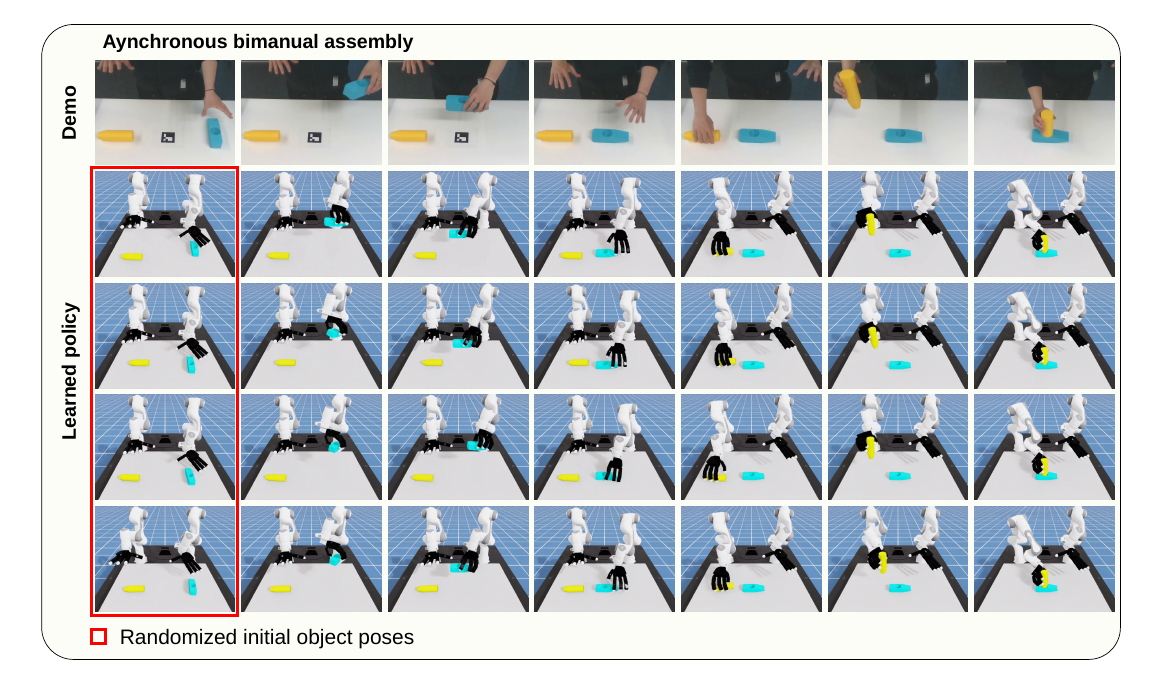}
\caption{Physics-based simulation of bimanual assembly skills learned from human videos for the asynchronous assembly task. The top row shows the collected demo and the rest rows show manipulation results with trained policy. We note that although our method is based on single demonstration, applying randomization on the initial poses of objects can still lead to generalization on random initial object poses. }
\label{fig:sim-exp-async}
\end{figure*}

\begin{figure*}[!t]
\centering
\includegraphics[trim=6mm 0mm 6mm 0mm,clip,width=1\linewidth]{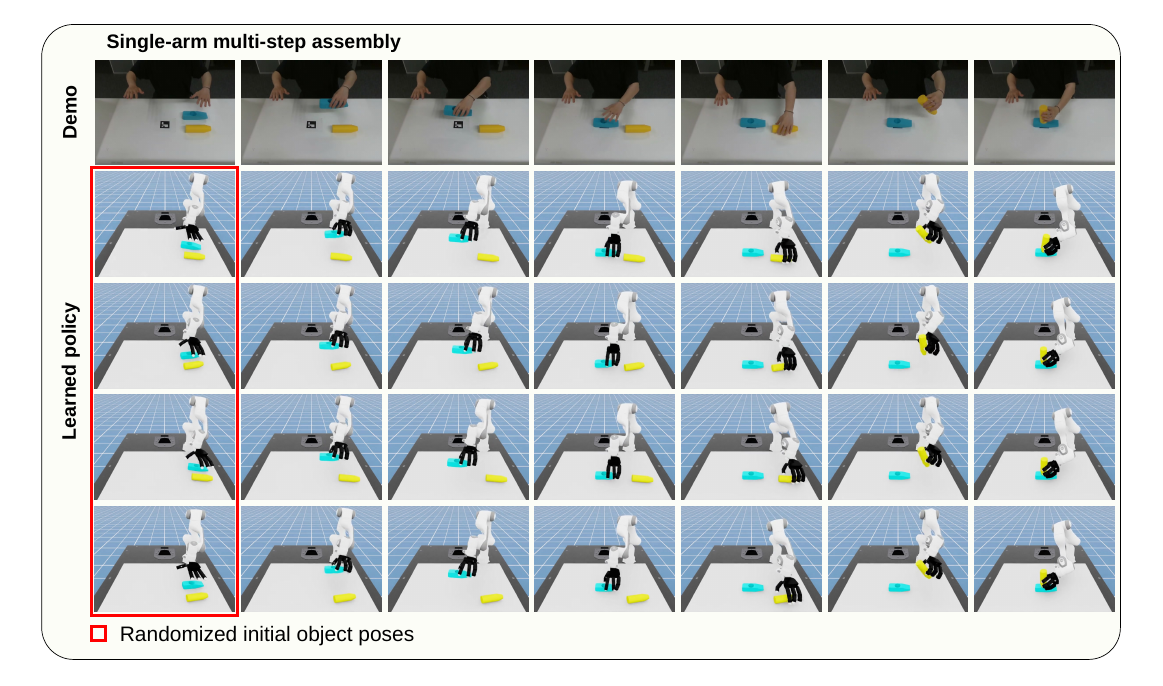}
\vspace{-8mm}
\caption{Physics-based simulation of bimanual assembly skills learned from human videos for the multi-step single arm assembly task. The top row shows the collected demo and the rest rows show manipulation results with trained policy. We note that although our method is based on single demonstration, applying randomization on the initial poses of objects can still lead to generalization on random initial object poses. }
\label{fig:sim-exp-single}
\vspace{-3mm}
\end{figure*}

\subsection{Experimental Setup}
\label{subsec:exp-setup}
We conduct extensive simulation experiments followed by a real-world proof-of-concept demonstration. Below we describe our settings for both simulation and real-world experiments. 
A more intuitive understanding of the task setup can be seen in the supplementary video. 

\textbf{Simulation Environment.}
We evaluate the key components of our method using the IsaacLab simulator~\cite{mittal2023orbit}, modelling a dual-arm setup composed of two Franka Panda arms, each equipped with an Allegro Hand. We evaluate both
a synchronous and an asynchronous bimanual assembly, to test different aspects of complex manipulation. 
The \textit{synchronous} setting (5 sub-goals) requires the two arms to cooperatively grasp a base and a peg and assemble them in mid-air. Its primary challenge is maintaining tight temporal and spatial coordination between two arms throughout the entire trajectory.
The \textit{asynchronous} task (11 sub-goals) has one arm placing the base before the second arm inserts the handle into the placed base. Its main challenge is navigating the significant distributional shift, or ``switch phase". 
Each training episode of these two settings adds random translational (in $[-10, 10]$ cm) and rotational (in $[-0.1, 0.1]$ radians) variation to the initial poses of the objects. 

\textbf{Real-World Validation.}
To demonstrate sim-to-real transferability and cross-embodiment capability, the proposed method is deployed on a physical system consisting of a UR-3e robotic arm and a XHand dexterous hand. 
As hardware constraints limit the real-world validation to a single-arm setup, the task is slightly reformulated into two variants: (1) \textit{One-step assembly}: The base object is at a fixed position on the table; the robot needs to grasp and insert the peg object to complete the assembly; (2) \textit{Multi-step place-and-assembly}: The robot needs to first pick and place the base object, then complete the assembly task by grasping and inserting the peg object. 
These experiments serve as a proof-of-concept, confirming that the proposed approach can be successfully transferred to real hardwares. 
The supplementary video shows additional qualitative results.

\subsection{Implementation Details\label{sec4b}}
\textbf{Data collection.} 
For demonstration recording, we use a single RealSense D455 camera to record synchronized RGB-D stream of human operation. We use 3D printed objects for data collection and their 3D models are later used for object pose estimation, refinement and also in the physical simulation. 

\textbf{Data processing module.} 
We use WiLoR~\cite{wilor} for hand pose estimation, Segmentation-and-Track Anything~\cite{cheng2023segment} for object mask segmentation, FoundationPose~\cite{wen2024foundationpose} for 3D object pose tracking and estimation in both the data processing module and real robot experiments. For the proposed assembly task(the robot inserts a peg-like object into a hole), the task-specific objective function $f_{task}$ is a weighted sum of two terms: (1) a peg-hole axis co-linearity loss and (2) a peg-hole endpoint relative position loss.

\textbf{Residual RL module.}
The object keypoints are used to represent the object state. For axis-symmetric objects like the cylinder, three points from its principle axis are used as keypoints, otherwise the rotated bounding box corners are used as keypoints.
We empirically found that $p_{init} \in [0.90, 0.95]$ is a good value for the success-gated reset strategy. The PPO~\cite{schulman2017proximal} algorithm from RSL\_RL library~\cite{rudin2022learning} is used to train our RL policy. The hyperparameters of PPO are determined by running parameter optimization with Optuna~\cite{akiba2019optuna}. The core hyperparameters of PPO we used in our experiments are list in~\ref{tab:rl-param}. 

\begin{wraptable}{r}{0.4\linewidth} 
\centering
\vspace{-4mm}
\caption{Core PPO hyperparameters}
\label{tab:rl-param}
\begin{tabular}{l|c}
\hline
\multicolumn{1}{c|}{\textbf{parameter}} & \textbf{value} \\ \hline
num\_steps\_per\_env          &24         \\
empirical\_normalization      &True       \\
value\_normalization          &True       \\
value\_loss\_coef             &0.5        \\
use\_clipped\_value\_loss     &True       \\
clip\_param                   &0.2        \\
entropy\_coef                 &5.0e-5     \\
num\_learning\_epochs         &4          \\
num\_mini\_batches            &32         \\
learning\_rate                &5.0e-4     \\
schedule                      &adaptive   \\
gamma                         &0.99       \\
lam                           &0.95       \\
desired\_kl                   &0.008      \\
max\_grad\_norm                &1.0        \\ \hline
\end{tabular}
\vspace{-5mm}
\end{wraptable}

\textbf{Physics-Based Actuation Dynamics for Sim-to-Real Transfer.}
To bridge the gap between non-linear electric motors on the real robot and the ideal actuator in simulation, we designed a randomized physics-based actuation model that embeds essential motor dynamics and systematic parameter variations directly into the low-level joint control loop in the simulation, for such randomization enables direct deployment of the RL policy on real robot. The default PD controller in the physics simulation uses an ideal torque source for simulating the joint-level position control, which is the conventional PD control in the simulation. In this case, the torque source is perfect and the behaviour of the simulated actuators do not represent detailed electrical properties and dynamics of real motors and its gear transmission system.

Unlike conventional way of simulating ideal PD control, our formulation explicitly accounts for 3 critical aspects of real actuators: (1) gain variations and calibration errors, (2) velocity-dependent torque saturation, and (3) sensor measurement biases. The actuator dynamics are expressed through physics-informed equations that capture fundamental motor constraints. 

The desired torque is computed as:
$$\tau_{\text{des}} = K_p^{\prime} \cdot (q_{\text{des}} - (\hat{q} + b)) + K_d^{\prime} \cdot (\dot{q}_{\text{des}} - \dot{q}),$$
where $K_p^{\prime} = \alpha_p K_p$ and $K_d^{\prime} = \alpha_d K_d$ denote randomized stiffness and damping gains with $\alpha_p, \alpha_d \in \mathcal{U}(0.9, 1.1)$. The term $b \in \mathcal{U}(-0.1, 0.1)$ radians models a constant position measurement bias, while $\hat{q}$ is the biased joint position. 

To capture realistic motor torque limitations, we incorporate a velocity-dependent saturation model reflecting the torque-speed characteristics of DC motors:
$$
\tau_{\text{max}}(\omega) = \frac{\tau_{\text{stall}}}{1 - \nu} \left(1 - \frac{|\omega|}{\omega_{\text{max}}}\right), \hspace{2mm} \tau_{\text{min}}(\omega) = \frac{\tau_{\text{stall}}}{1 - \nu} \left(-1 - \frac{|\omega|}{\omega_{\text{max}}}\right)
$$
where $\tau_{\text{stall}}$ is the stall torque, $\omega$ is the joint angular velocity, and $\omega_{\text{max}}$ the no-load angular velocity. The parameter $\nu \in \mathcal{U}(\tfrac{1}{3}, \tfrac{2}{3})$ specifies the normalized velocity at which torque saturation begins, capturing variations in motor constants and electrical characteristics across actuators. The final applied torque is calculated as $\tau_{\text{applied}} = \gamma \cdot \text{clip}(\tau_{\text{des}}, \tau_{\text{min}}(\omega), \tau_{\text{max}}(\omega))$
where $\gamma \in \mathcal{U}(0.9, 1.1)$ introduces overall motor strength variability, capturing differences in motor constants and amplifier gains. 

During training, all randomization parameters ($\alpha_p, \alpha_d, b, \nu, \gamma$) are independently resampled for each joint at the beginning of every episode.

\subsection{Experiment in Simulation and Ablation Study} 
\begin{wraptable}{r}{0.5\linewidth} 
\centering
\caption{Number of sub-goals reached with only extracted motion priors, RL trained from scratch and our proposed RL pipeline (motion priors + RL). }
\label{tab:my-table}
\resizebox{\linewidth}{!}{%
\begin{tabular}{c|cc}
\hline
 & \textbf{Synchronous task} & \textbf{Asynchronous task} \\ \hline
Motion prior-only & 0/5 &  0/11 \\
RL-only & 0/5 & 1/11 \\ \hline
Motion prior+RL (\textbf{Ours}) & \textbf{5/5} & \textbf{11/11} \\ \hline
\end{tabular}%
}
\end{wraptable}
This section presents the simulation's experimental results and evaluates the effectiveness of the key design choices, including: 
(1) \textit{Motion Prior}: comparing RL based on demonstration guidance against learning from scratch;
(2) \textit{Residual Action Clipping}: examining the importance of constraining the policy to remain close to the reference trajectory for training stability;
(3) \textit{Pre-grasp Curriculum}: assessing whether our dedicated reaching stage contributes to more robust and stable grasps;
(4) \textit{Temporal-segment based RL}: comparing the effect of segmenting the task versus learning from the full trajectory on long-horizon performance;
(5) \textit{Success-Gated Reset Strategy}: analyzing the effect of the proposed reset mechanism on exploration efficiency and training stability.
Snapshots of synchronous and asynchronous assembly task and multi-step single-arm assembly task in simulation are presented in Figure.~\ref{fig:sim-exp-sync},~\ref{fig:sim-exp-async},~\ref{fig:sim-exp-single} respectively. 

\begin{figure}[!t]
\centering
\includegraphics[trim=5mm 2mm 5mm 2mm,clip,width=1.0\linewidth]{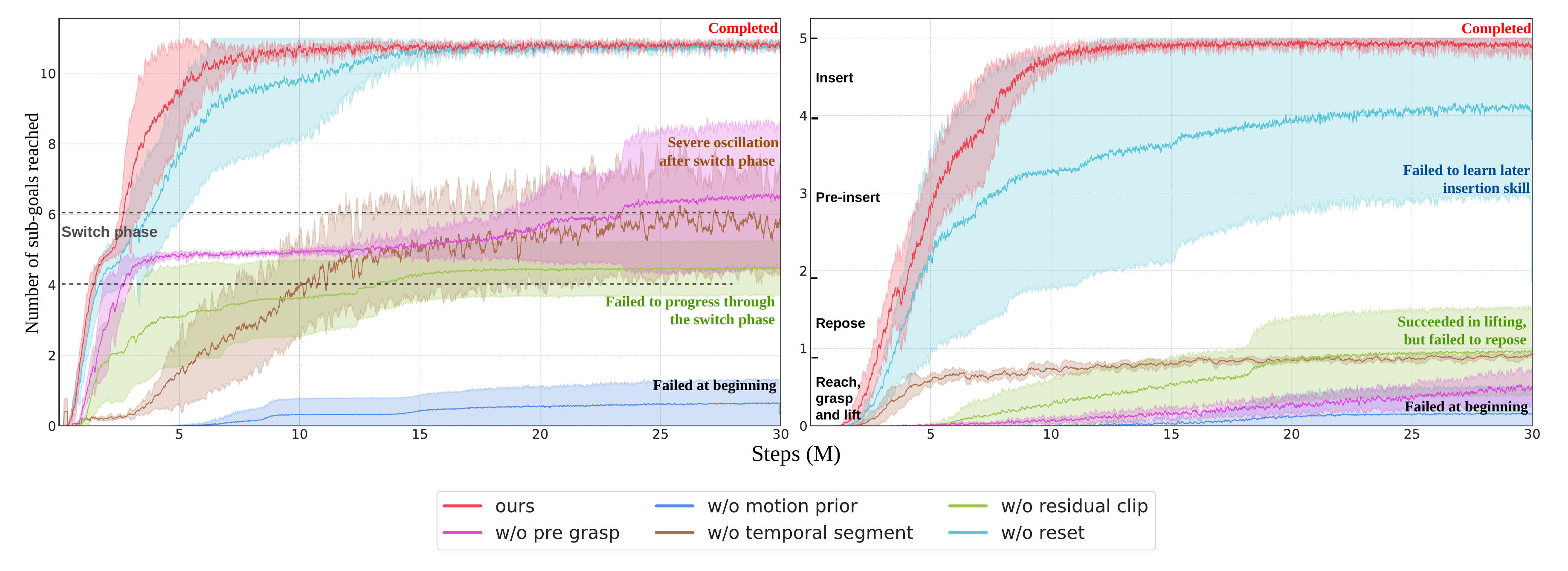}
\caption{Quantitative comparison and ablation results on the asynchronous (\textit{left}) and synchronous (\textit{right}) bimanual assembly tasks. The plots show the average number of achieved sub-goals (vertical) versus number of training steps, for different ablated versions of the proposed algorithm.} 
\label{fig:ablation-comparison}
\end{figure}

We perform ablation studies on both synchronous and asynchronous assembly tasks. All experiments were repeated five times with different random seeds.
The results in Table.~\ref{tab:my-table} show that neither by replaying the extracted motion priors nor by learning RL agent from scratch can complete the task. The corresponding learning curves are presented in Figure.~\ref{fig:ablation-comparison}. The solid curves are the mean values across all runs while the shaded area denotes the standard deviations. The full method (\textcolor{wandb-red}{red}) achieves the highest performance and sample efficiency across both tasks, successfully accomplishing all sub-goals. The following is a detailed analysis of each component that contributes to this overall best performance.

\textbf{Effectiveness of Demo-guided and Temporal-segment based RL.} Either learning from scratch (curves \textcolor{wandb-blue}{w/o motion prior}) or from a single, monolithic trajectory (curves \textcolor{wandb-brown}{w/o temporal segment}) results in a complete failure on both tasks. This is attributed to the high-dimensional state-action space, where random exploration is intractable. Furthermore, learning from an entire long trajectory leads to severe temporal misalignment; the agent receives meaningless base actions from a different task phase, polluting the learning signal and preventing progress. These results confirm that our core approach of using a segmented demonstration as a motion prior is essential.

\textbf{Effectiveness of Residual Action Clipping.} Disabling the residual clip (curve \textcolor{wandb-green}{w/o residual clip}) degrades performance on both tasks by allowing the agent to drift too far from the stable motion prior. The effect is more pronounced on the synchronous task, where the policy plateaus at a much lower performance. The delicate coordination required has a very small margin for error, which is easily disrupted by the large, erratic actions from unconstrained exploration.

\textbf{Effectiveness of the Pre-Grasp Curriculum.} Removing the pre-grasp (curve \textcolor{wandb-purple}{w/o pre grasp}) is catastrophic for the synchronous task.
A stable, simultaneous grasp by both hands is a hard prerequisite for success in the synchronous task; learning arm and finger control together from the start leads to unstable reaching behavior and immediate failure. In the asynchronous task, while learning is significantly hindered initially, the agent eventually recovers. This suggests that while the pre-grasp curriculum is always beneficial, it is most critical in tasks with tight coordination constraints.

\textbf{Effectiveness of the Success-Gated Resets.} Removing the reset strategy (curve \textcolor{wandb-cyan}{w/o reset}) reveals a critical bottleneck in long-horizon learning. While the agent in the asynchronous task eventually succeeds, its learning rate for the post-switch phase is significantly degraded.
The more challenging synchronous task 
has hard failure, as the agent plateaus early and never masters the final coordination stages. The stricter success conditions of the synchronous task amplify the data imbalance problem; without the reset strategy to provide focused practice, the agent cannot learn from the handful of times it may randomly reach these difficult states. Thus, success-gated resets are crucial for efficient exploration and mastery of complex, sequential tasks.

\subsection{Real-world Experiments}

\begin{figure*}[t]
\centering
\includegraphics[trim=6mm 0mm 6mm 0mm,clip,width=1\linewidth]{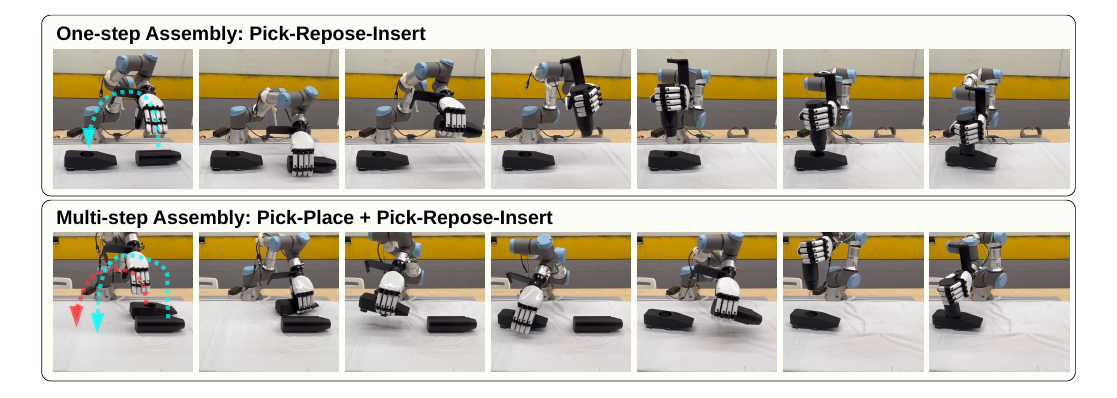}
\vspace{-5mm}
\caption{Real-robot experiments: (\textit{top}) One-step assembly with a sequential manipulation of pick-repose-insert; (\textit{bottom}) Multi-step assembly with a sequential manipulation of pick-place of the base object followed by pick-repose-insert of another. }
\vspace{-3mm}
\label{fig:GRASP}
\end{figure*}

To validate the sim-to-real transferability and cross-embodiment capability of the proposed framework, we conduct real-world experiments on the task described in Section.~\ref{subsec:exp-setup}. We collect and process the single demonstration with exactly the same pipeline as used for the simulation experiments, but retarget and replay the collected demonstration on the real-world robot setup~(UR-3e + XHand). During policy training, to bridge the gap between real motors and the ideal ones in simulation, we apply extra physical-based actuation randomization as detailed in Section.~\ref{sec4b}. 

We conducted 20 trials of the \textit{One-step assembly} task with varying initial object states to evaluate the success rate of the trained agent. Snapshots of these experiments are presented in Fig. \ref{fig:GRASP}. 
The agent succeeded in 18 out of 20 runs, achieving a 90\% success rate. 
while in the \textit{Multi-step assembly} task, the agent succeeded in 12 runs, resulting in a 60\% success rate. 
In the real-world experiments, the failures are mainly due to the drift of the object pose estimates when the object is occluded by the robot hand. 
This estimation error has a more significant impact on the \textit{Multi-step assembly} task, where a misplacement of the base object will hugely affect the result of later insertion stage. 
The estimation errors results in erroneous state input into the RL agent, making the RL agent unable to reposition the object to the correct insertion pose. 
Overall, the real-world experiments successfully demonstrate the sim-to-real transferability and cross-embodiment capability of the proposed DemoBot framework, highlighting its contribution to the robotic community.

\section{Discussion}
\label{sec:conclusion}
This technical report presented DemoBot, a novel framework for efficiently learning of complex dexterous bimanual manipulation skills from a single RGB-D video. 
The main contributions include a comprehensive data processing module which transfers the unstructed video into structured motion priors for robot learning, and a novel corrective residual RL pipeline with a set of novel designs for addressing the challenges in learning long-horizon dexterous bimanual manipulation skills. While the method has significant advantages, its limitations include the reliance on manually selected keyframes for task decomposition, depth information and 3D printed objects for object pose estimation and a manually specified task objective for object pose refinement. 
However, we claim that the approach is a promising way of learning complex manipulation skills from unstructured human videos. Our future work will focus on addressing these limitations and exploring how to scale up the approach to more diverse and even imperfect demonstrations, progressing towards scalable manipulation skill learning from in-the-wild videos. 

\section{Future work}
Although the proposed DemoBot steps a significant step towards learning dexterous manipulation skills from raw human videos, several core relaxation have been made to simplify this challenging research problem: (1) RGB-D camera is used to capture both RGB and depth information during demonstration; (2) Existing 3D object models are used for object pose estimation and for policy learning in simulation; (3) Manual annotation is required to select keyframes from the raw video; (4) Task-specific optimization is needed to refine the raw estimated object poses. Additionally, the 6D pose is currently used to represent the state of object, however, this representation is limited to rigid object only. 

Future work would need a community effort to eliminate these relaxations. Combining depth estimation models~\cite{depth_anything_v1, depth_anything_v2} and 3D assets generation models~\cite{lai2025hunyuan3d25highfidelity3d, hunyuan3d22025tencent} would have the potential to extend the current framework from RGB-D video to arbitrary RGB-only videos and get rid of the dependence of the existing 3D object models, eliminating the relaxation (1) and (2). Vision-Language Models~(VLMs) could be used to understand the sequential task visually and select the keyframes corresponding to each task phase. Moreover, recent work~\cite{ma2023eureka} shows that the Large-Language Models~(LLMs) have the potential to automatically generate reward functions for RL training. Therefore, the relaxation (3) and (4) could be eliminated by incorporating VLMs into the current framework. 

Furthermore, future works can focus on extend the current framework to more diverse type of the manipulated objects, e.g. articulated objects, deformable objects, by adopting a more generalized representation of the object state. For example, a combination of 6D pose and keypoints, 6D pose can be used to roughly represent the current pose of the entire object, while the keypoints represent the deformation or the articulation of the object. 

\section{Conclusion}
In summary, this study serves as a pioneer for developing an effective framework for learning bimanual dexterous manipulation skills from arbitrary internet video, which is validated through extensive experiments in both simulation and real-world setup. Such a framework demonstrates its effectiveness at a system-level of engineering and integration, using techniques ranging from visual-based hand reconstruction, object pose estimation, to a novel demo-augmented residual reinforcement learning. 

Moreover, this work presents a significant step toward scalable robot learning. By extracting motion priors from a single, imperfect human video, DemoBot bypasses the data bottleneck of teleoperation. Our results demonstrate that combining suboptimal motion priors with residual RL enables robots to master contact-rich, long-horizon tasks -- problems that are intractable for RL from scratch due to exploration complexity, yet too costly for imitation learning due to data requirements. This paradigm opens the door to leveraging internet-scale video data, moving us closer to general-purpose robots capable of acquiring new dexterous skills by simply `watching' humans.

\newpage
\section{Contributions \& Acknowledgements}
\label{sec:contributions}
The following contributions are categorized by technical and research inputs, and the authors' names are listed within each category.

\begin{itemize}
    \item Research Ideas and Conceptualization: Yucheng Xu, Robert B. Fisher, Zhibin Li.
    \item Data collection and processing module: Yucheng Xu.
    \item Residual RL module: Yucheng Xu, Xiaofeng Mao, Elle Miller.
    \item Experiments in simulation: Yucheng Xu, Xiaofeng Mao Elle Miller.
    \item Experiments on real robot: Xinyu Yi, Yang Li, Zhibin Li.
    \item Hardware \& Software Infrastructure: Yucheng Xu, Xinyu Yi, Yang Li.  
    \item Paper Writing: All authors.
    \item Research Direction and Team Lead: Robert B. Fisher, Zhibin Li.
\end{itemize}  
We sincerely thank Jianran Liu, Yufei Liu, Lisuo Li for their strong support on hardware setup. We extend our sincere gratitude to Wenjia Zhu for encouraging a focus on high‑impact research. We are also deeply thankful to Yonghui Wu for his strategic vision in shaping our research priorities. In addition, we gratefully acknowledge the Department Head, Hang Li, for his invaluable organizational support throughout this project.

\clearpage

\bibliographystyle{plainnat}
\bibliography{main}

\begin{thebibliography}{45}
\providecommand{\natexlab}[1]{#1}
\providecommand{\url}[1]{\texttt{#1}}
\expandafter\ifx\csname urlstyle\endcsname\relax
  \providecommand{\doi}[1]{doi: #1}\else
  \providecommand{\doi}{doi: \begingroup \urlstyle{rm}\Url}\fi

\bibitem[Akiba et~al.(2019)Akiba, Sano, Yanase, Ohta, and Koyama]{akiba2019optuna}
Takuya Akiba, Shotaro Sano, Toshihiko Yanase, Takeru Ohta, and Masanori Koyama.
\newblock {O}ptuna: A next-generation hyperparameter optimization framework.
\newblock In \emph{The 25th ACM SIGKDD International Conference on Knowledge Discovery \& Data Mining}, pages 2623--2631, 2019.

\bibitem[Chen et~al.(2023)Chen, Geng, Zhong, Ji, Jiang, Lu, Dong, and Yang]{chen2023bi}
Yuanpei Chen, Yiran Geng, Fangwei Zhong, Jiaming Ji, Jiechuang Jiang, Zongqing Lu, Hao Dong, and Yaodong Yang.
\newblock Bi-dexhands: Towards human-level bimanual dexterous manipulation.
\newblock \emph{IEEE Transactions on Pattern Analysis and Machine Intelligence}, 46\penalty0 (5):\penalty0 2804--2818, 2023.

\bibitem[Chen et~al.(2024)Chen, Wang, Yang, and Liu]{chen2024object}
Yuanpei Chen, Chen Wang, Yaodong Yang, and C~Karen Liu.
\newblock Object-centric dexterous manipulation from human motion data.
\newblock \emph{arXiv:2411.04005}, 2024.

\bibitem[Cheng et~al.(2024)Cheng, Li, Yang, Yang, and Wang]{cheng2024open}
Xuxin Cheng, Jialong Li, Shiqi Yang, Ge~Yang, and Xiaolong Wang.
\newblock Open-television: Teleoperation with immersive active visual feedback.
\newblock \emph{arXiv:2407.01512}, 2024.

\bibitem[Cheng et~al.(2023)Cheng, Li, Xu, Li, Yang, Wang, and Yang]{cheng2023segment}
Yangming Cheng, Liulei Li, Yuanyou Xu, Xiaodi Li, Zongxin Yang, Wenguan Wang, and Yi~Yang.
\newblock Segment and track anything.
\newblock \emph{arXiv:2305.06558}, 2023.

\bibitem[Ding et~al.(2024)Ding, Qin, Zhu, Jia, Yang, Yang, Qi, and Wang]{ding2024bunny}
Runyu Ding, Yuzhe Qin, Jiyue Zhu, Chengzhe Jia, Shiqi Yang, Ruihan Yang, Xiaojuan Qi, and Xiaolong Wang.
\newblock Bunny-visionpro: Real-time bimanual dexterous teleoperation for imitation learning.
\newblock \emph{arXiv:2407.03162}, 2024.

\bibitem[Gupta et~al.(2019)Gupta, Kumar, Lynch, Levine, and Hausman]{gupta2019relay}
Abhishek Gupta, Vikash Kumar, Corey Lynch, Sergey Levine, and Karol Hausman.
\newblock Relay policy learning: Solving long-horizon tasks via imitation and reinforcement learning.
\newblock \emph{arXiv preprint arXiv:1910.11956}, 2019.

\bibitem[Handa et~al.(2020)Handa, Van~Wyk, Yang, Liang, Chao, Wan, Birchfield, Ratliff, and Fox]{handa2020dexpilot}
Ankur Handa, Karl Van~Wyk, Wei Yang, Jacky Liang, Yu-Wei Chao, Qian Wan, Stan Birchfield, Nathan Ratliff, and Dieter Fox.
\newblock Dexpilot: Vision-based teleoperation of dexterous robotic hand-arm system.
\newblock In \emph{2020 IEEE Int Conf on Robotics and Automation (ICRA)}, pages 9164--9170, 2020.

\bibitem[Hasson et~al.(2019)Hasson, Varol, Tzionas, Kalevatykh, Black, Laptev, and Schmid]{hasson19_obman}
Yana Hasson, G{\"u}l Varol, Dimitris Tzionas, Igor Kalevatykh, Michael~J. Black, Ivan Laptev, and Cordelia Schmid.
\newblock Learning joint reconstruction of hands and manipulated objects.
\newblock In \emph{CVPR}, 2019.

\bibitem[Hester et~al.(2018)Hester, Vecerik, Pietquin, Lanctot, Schaul, Piot, Horgan, Quan, Sendonaris, Osband, et~al.]{hester2018deep}
Todd Hester, Matej Vecerik, Olivier Pietquin, Marc Lanctot, Tom Schaul, Bilal Piot, Dan Horgan, John Quan, Andrew Sendonaris, Ian Osband, et~al.
\newblock {Deep Q-learning from Demonstrations}.
\newblock In \emph{the AAAI conference on artificial intelligence}, volume~32, 2018.

\bibitem[Ho and Ermon(2016)]{ho2016generative}
Jonathan Ho and Stefano Ermon.
\newblock Generative adversarial imitation learning.
\newblock \emph{Advances in neural information processing systems}, 29, 2016.

\bibitem[Huey et~al.(2025)Huey, Wang, Wu, Artzi, and Choudhury]{huey2025imitation}
William Huey, Huaxiaoyue Wang, Anne Wu, Yoav Artzi, and Sanjiban Choudhury.
\newblock Imitation learning from a single temporally misaligned video.
\newblock \emph{arXiv preprint arXiv:2502.05397}, 2025.

\bibitem[Johannink et~al.(2019)Johannink, Bahl, Nair, Luo, Kumar, Loskyll, Ojea, Solowjow, and Levine]{johannink2019residual}
Tobias Johannink, Shikhar Bahl, Ashvin Nair, Jianlan Luo, Avinash Kumar, Matthias Loskyll, Juan~Aparicio Ojea, Eugen Solowjow, and Sergey Levine.
\newblock Residual reinforcement learning for robot control.
\newblock In \emph{2019 international conference on robotics and automation (ICRA)}, pages 6023--6029. IEEE, 2019.

\bibitem[Li et~al.(2025)Li, Li, Liu, Li, and Huang]{li2025maniptrans}
Kailin Li, Puhao Li, Tengyu Liu, Yuyang Li, and Siyuan Huang.
\newblock Maniptrans: Efficient dexterous bimanual manipulation transfer via residual learning.
\newblock In \emph{Proceedings of the Computer Vision and Pattern Recognition Conference}, pages 6991--7003, 2025.

\bibitem[Li et~al.(2019)Li, Ma, Liang, G{\"o}rner, Ruppel, Fang, Sun, and Zhang]{li2019vision}
Shuang Li, Xiaojian Ma, Hongzhuo Liang, Michael G{\"o}rner, Philipp Ruppel, Bin Fang, Fuchun Sun, and Jianwei Zhang.
\newblock Vision-based teleoperation of shadow dexterous hand using end-to-end deep neural network.
\newblock In \emph{2019 International Conference on Robotics and Automation (ICRA)}, pages 416--422. IEEE, 2019.

\bibitem[Lin et~al.(2021)Lin, Wang, and Liu]{metro}
Kevin Lin, Lijuan Wang, and Zicheng Liu.
\newblock End-to-end human pose and mesh reconstruction with transformers.
\newblock In \emph{CVPR}, 2021.

\bibitem[Ma et~al.(2023)Ma, Liang, Wang, Huang, Bastani, Jayaraman, Zhu, Fan, and Anandkumar]{ma2023eureka}
Yecheng~Jason Ma, William Liang, Guanzhi Wang, De-An Huang, Osbert Bastani, Dinesh Jayaraman, Yuke Zhu, Linxi Fan, and Anima Anandkumar.
\newblock Eureka: Human-level reward design via coding large language models.
\newblock \emph{arXiv preprint arXiv:2310.12931}, 2023.

\bibitem[Mao et~al.(2023)Mao, Xu, Wen, Kasaei, Yu, Psomopoulou, Lepora, and Li]{mao2023learning}
Xiaofeng Mao, Yucheng Xu, Ruoshi Wen, Mohammadreza Kasaei, Wanming Yu, Efi Psomopoulou, Nathan~F Lepora, and Zhibin Li.
\newblock Learning fine pinch-grasp skills using tactile sensing from real demonstration data.
\newblock \emph{CoRR}, 2023.

\bibitem[Mao et~al.(2024)Mao, Giudici, Coppola, Althoefer, Farkhatdinov, Li, and Jamone]{mao2024dexskills}
Xiaofeng Mao, Gabriele Giudici, Claudio Coppola, Kaspar Althoefer, Ildar Farkhatdinov, Zhibin Li, and Lorenzo Jamone.
\newblock Dexskills: Skill segmentation using haptic data for learning autonomous long-horizon robotic manipulation tasks.
\newblock In \emph{IEEE/RSJ International Conference on Intelligent Robots and Systems}, 2024.

\bibitem[McCarthy et~al.(2025)McCarthy, Tan, Schmidt, Acero, Herr, Du, Thuruthel, and Li]{mccarthy2025towards}
Robert McCarthy, Daniel~CH Tan, Dominik Schmidt, Fernando Acero, Nathan Herr, Yilun Du, Thomas~G Thuruthel, and Zhibin Li.
\newblock Towards generalist robot learning from internet video: A survey.
\newblock \emph{J. Artificial Intelligence Research}, 83, 2025.

\bibitem[Mittal et~al.(2023)Mittal, Yu, Yu, Liu, Rudin, Hoeller, Yuan, Singh, Guo, Mazhar, Mandlekar, Babich, State, Hutter, and Garg]{mittal2023orbit}
Mayank Mittal, Calvin Yu, Qinxi Yu, Jingzhou Liu, Nikita Rudin, David Hoeller, Jia~Lin Yuan, Ritvik Singh, Yunrong Guo, Hammad Mazhar, Ajay Mandlekar, Buck Babich, Gavriel State, Marco Hutter, and Animesh Garg.
\newblock Orbit: A unified simulation framework for interactive robot learning environments.
\newblock \emph{IEEE Robotics and Automation Letters}, 8\penalty0 (6):\penalty0 3740--3747, 2023.
\newblock \doi{10.1109/LRA.2023.3270034}.

\bibitem[Ni et~al.(2023)Ni, Ma, Eysenbach, and Bacon]{ni2023transformers}
Tianwei Ni, Michel Ma, Benjamin Eysenbach, and Pierre-Luc Bacon.
\newblock When do transformers shine in rl? decoupling memory from credit assignment.
\newblock \emph{Advances in Neural Information Processing Systems}, 36:\penalty0 50429--50452, 2023.

\bibitem[Pavlakos et~al.(2024)Pavlakos, Shan, Radosavovic, Kanazawa, Fouhey, and Malik]{hamer}
Georgios Pavlakos, Dandan Shan, Ilija Radosavovic, Angjoo Kanazawa, David Fouhey, and Jitendra Malik.
\newblock Reconstructing hands in 3{D} with transformers.
\newblock In \emph{CVPR}, 2024.

\bibitem[Potamias et~al.(2025)Potamias, Zhang, Deng, and Zafeiriou]{wilor}
Rolandos~Alexandros Potamias, Jinglei Zhang, Jiankang Deng, and Stefanos Zafeiriou.
\newblock Wilor: End-to-end 3d hand localization and reconstruction in-the-wild.
\newblock In \emph{Proc Computer Vision and Pattern Recognition}, pages 12242--12254, 2025.

\bibitem[Qin et~al.(2021)Qin, Wu, Liu, Jiang, Yang, Fu, and Wang]{qin2021dexmv}
Yuzhe Qin, Yueh-Hua Wu, Shaowei Liu, Hanwen Jiang, Ruihan Yang, Yang Fu, and Xiaolong Wang.
\newblock Dexmv: Imitation learning for dexterous manipulation from human videos, 2021.

\bibitem[Qin et~al.(2022)Qin, Su, and Wang]{qin2022one}
Yuzhe Qin, Hao Su, and Xiaolong Wang.
\newblock From one hand to multiple hands: Imitation learning for dexterous manipulation from single-camera teleoperation.
\newblock \emph{IEEE Robotics and Automation Letters}, 7\penalty0 (4):\penalty0 10873--10881, 2022.

\bibitem[Qin et~al.(2023)Qin, Yang, Huang, Van~Wyk, Su, Wang, Chao, and Fox]{qin2023anyteleop}
Yuzhe Qin, Wei Yang, Binghao Huang, Karl Van~Wyk, Hao Su, Xiaolong Wang, Yu-Wei Chao, and Dieter Fox.
\newblock Anyteleop: A general vision-based dexterous robot arm-hand teleoperation system.
\newblock In \emph{Robotics: Science \& Systems}, 2023.

\bibitem[Rajeswaran et~al.(2017)Rajeswaran, Kumar, Gupta, Vezzani, Schulman, Todorov, and Levine]{rajeswaran2017learning}
Aravind Rajeswaran, Vikash Kumar, Abhishek Gupta, Giulia Vezzani, John Schulman, Emanuel Todorov, and Sergey Levine.
\newblock Learning complex dexterous manipulation with deep reinforcement learning and demonstrations.
\newblock \emph{arXiv:1709.10087}, 2017.

\bibitem[Romero et~al.(2017)Romero, Tzionas, and Black]{MANO}
Javier Romero, Dimitrios Tzionas, and Michael~J. Black.
\newblock Embodied hands: Modeling and capturing hands and bodies together.
\newblock \emph{ACM Transactions on Graphics, (Proc. SIGGRAPH Asia)}, 36\penalty0 (6), November 2017.

\bibitem[Rudin et~al.(2022)Rudin, Hoeller, Reist, and Hutter]{rudin2022learning}
Nikita Rudin, David Hoeller, Philipp Reist, and Marco Hutter.
\newblock Learning to walk in minutes using massively parallel deep reinforcement learning.
\newblock In \emph{Proceedings of the 5th Conference on Robot Learning}, volume 164 of \emph{Proceedings of Machine Learning Research}, pages 91--100. PMLR, 2022.
\newblock URL \url{https://proceedings.mlr.press/v164/rudin22a.html}.

\bibitem[Schulman et~al.(2017)Schulman, Wolski, Dhariwal, Radford, and Klimov]{schulman2017proximal}
John Schulman, Filip Wolski, Prafulla Dhariwal, Alec Radford, and Oleg Klimov.
\newblock Proximal policy optimization algorithms.
\newblock \emph{arXiv preprint arXiv:1707.06347}, 2017.

\bibitem[Shaw et~al.(2024)Shaw, Li, Yang, Srirama, Liu, Xiong, Mendonca, and Pathak]{shaw2024bimanual}
Kenneth Shaw, Yulong Li, Jiahui Yang, Mohan~Kumar Srirama, Ray Liu, Haoyu Xiong, Russell Mendonca, and Deepak Pathak.
\newblock Bimanual dexterity for complex tasks.
\newblock \emph{arXiv:2411.13677}, 2024.

\bibitem[Silver et~al.(2018)Silver, Allen, Tenenbaum, and Kaelbling]{silver2018residual}
Tom Silver, Kelsey Allen, Josh Tenenbaum, and Leslie Kaelbling.
\newblock Residual policy learning.
\newblock \emph{arXiv:1812.06298}, 2018.

\bibitem[Tao et~al.(2025)Tao, Srirama, Liu, Shaw, and Pathak]{tao2025dexwild}
Tony Tao, Mohan~Kumar Srirama, Jason~Jingzhou Liu, Kenneth Shaw, and Deepak Pathak.
\newblock Dexwild: Dexterous human interactions for in-the-wild robot policies.
\newblock \emph{arXiv:2505.07813}, 2025.

\bibitem[Team(2025{\natexlab{a}})]{hunyuan3d22025tencent}
Tencent~Hunyuan3D Team.
\newblock Hunyuan3d 2.0: Scaling diffusion models for high resolution textured 3d assets generation, 2025{\natexlab{a}}.

\bibitem[Team(2025{\natexlab{b}})]{lai2025hunyuan3d25highfidelity3d}
Tencent~Hunyuan3D Team.
\newblock Hunyuan3d 2.5: Towards high-fidelity 3d assets generation with ultimate details, 2025{\natexlab{b}}.
\newblock URL \url{https://arxiv.org/abs/2506.16504}.

\bibitem[Triantafyllidis et~al.(2023)Triantafyllidis, Acero, Liu, and Li]{triantafyllidis2023hybrid}
Eleftherios Triantafyllidis, Fernando Acero, Zhaocheng Liu, and Zhibin Li.
\newblock Hybrid hierarchical learning for solving complex sequential tasks using the robotic manipulation network roman.
\newblock \emph{Nature Machine Intelligence}, 5\penalty0 (9):\penalty0 991--1005, 2023.

\bibitem[Vecerik et~al.(2017)Vecerik, Hester, Scholz, Wang, Pietquin, Piot, Heess, Roth{\"o}rl, Lampe, and Riedmiller]{vecerik2017leveraging}
Mel Vecerik, Todd Hester, Jonathan Scholz, Fumin Wang, Olivier Pietquin, Bilal Piot, Nicolas Heess, Thomas Roth{\"o}rl, Thomas Lampe, and Martin Riedmiller.
\newblock Leveraging demonstrations for deep reinforcement learning on robotics problems with sparse rewards.
\newblock \emph{arXiv:1707.08817}, 2017.

\bibitem[Wang et~al.(2024)Wang, Shi, Wang, Zhang, Fei-Fei, and Liu]{wang2024dexcap}
Chen Wang, Haochen Shi, Weizhuo Wang, Ruohan Zhang, Li~Fei-Fei, and C~Karen Liu.
\newblock Dexcap: Scalable and portable mocap data collection system for dexterous manipulation.
\newblock \emph{arXiv:2403.07788}, 2024.

\bibitem[Wen et~al.(2024)Wen, Yang, Kautz, and Birchfield]{wen2024foundationpose}
Bowen Wen, Wei Yang, Jan Kautz, and Stan Birchfield.
\newblock Foundationpose: Unified 6d pose estimation and tracking of novel objects.
\newblock In \emph{Proceedings of the IEEE/CVF Conference on Computer Vision and Pattern Recognition}, pages 17868--17879, 2024.

\bibitem[Xiong et~al.(2022)Xiong, Fu, Zhang, Bao, Zhang, Huang, Xu, Garg, and Lu]{xiong2022robotube}
Haoyu Xiong, Haoyuan Fu, Jieyi Zhang, Chen Bao, Qiang Zhang, Yongxi Huang, Wenqiang Xu, Animesh Garg, and Cewu Lu.
\newblock Robotube: Learning household manipulation from human videos with simulated twin environments.
\newblock In \emph{6th Annual Conference on Robot Learning}, 2022.

\bibitem[Yang et~al.(2024{\natexlab{a}})Yang, Kang, Huang, Xu, Feng, and Zhao]{depth_anything_v1}
Lihe Yang, Bingyi Kang, Zilong Huang, Xiaogang Xu, Jiashi Feng, and Hengshuang Zhao.
\newblock Depth anything: Unleashing the power of large-scale unlabeled data.
\newblock In \emph{CVPR}, 2024{\natexlab{a}}.

\bibitem[Yang et~al.(2024{\natexlab{b}})Yang, Kang, Huang, Zhao, Xu, Feng, and Zhao]{depth_anything_v2}
Lihe Yang, Bingyi Kang, Zilong Huang, Zhen Zhao, Xiaogang Xu, Jiashi Feng, and Hengshuang Zhao.
\newblock Depth anything v2.
\newblock \emph{arXiv:2406.09414}, 2024{\natexlab{b}}.

\bibitem[Zhang et~al.(2025)Zhang, Hu, Yuan, and Xu]{zhang2025doglove}
Han Zhang, Songbo Hu, Zhecheng Yuan, and Huazhe Xu.
\newblock Doglove: Dexterous manipulation with a low-cost open-source haptic force feedback glove.
\newblock \emph{arXiv:2502.07730}, 2025.

\bibitem[Zhou et~al.(2025)Zhou, Wang, Tai, Deng, Liu, and Jia]{zhou2025you}
Huayi Zhou, Ruixiang Wang, Yunxin Tai, Yueci Deng, Guiliang Liu, and Kui Jia.
\newblock You only teach once: Learn one-shot bimanual robotic manipulation from video demonstrations.
\newblock \emph{arXiv:2501.14208}, 2025.

\end{thebibliography}




\end{document}